\documentclass[11pt]{article}

\usepackage[margin=1in]{geometry}
\usepackage{authblk}
\usepackage{amsmath,amssymb}
\usepackage{graphicx}
\usepackage{booktabs}
\usepackage{multirow}
\usepackage{array}
\usepackage{enumitem}
\usepackage{xcolor}
\usepackage{microtype}
\usepackage[numbers,sort&compress]{natbib}
\usepackage[hidelinks]{hyperref}
\usepackage{xurl}
\usepackage{url}

\title{\textbf{SkNeXt enables topology-guided neuronal reconstruction from petabyte-scale microscopy data}}

\author[1,2]{Jiayi Ding}
\author[1]{Hu Zhao}
\affil[1]{Chinese Institute for Brain Research, Beijing, China}
\affil[2]{Academy for Advanced Interdisciplinary Studies, Peking University, Beijing, China}
\date{September 2026}

\begin{document}
\maketitle

\begin{abstract}
Recent advances in high-resolution fluorescence and electron microscopy have enabled nanoscale imaging across increasingly large brain volumes, but the resulting terabyte- to petabyte-scale datasets make complete neuronal reconstruction prohibitively expensive in computation, data movement, and manual proofreading.
Here, we present SkNeXt, a topology-first framework for scalable neuronal reconstruction from large volumetric microscopy datasets. Instead of densely processing entire image volumes, SkNeXt first converts neuronal morphology into compact SWC skeletons that preserve long-range connectivity. Proofreading is therefore focused on sparse neuronal trees, allowing branch, continuity, and connectivity errors to be corrected before high-resolution reconstruction. The corrected skeletons then serve as persistent structural priors for recovering detailed morphology while preserving neuronal identity and topology.
Crucially, SkNeXt also uses neuronal skeletons as spatial indices for selective data access, retrieving high-resolution image regions only along reconstructed trajectories and bypassing most background and signal-free volumes. This substantially reduces I/O and computational overhead, allowing reconstruction cost to scale with neuronal morphology rather than total dataset size.
Using SkNeXt, we reconstructed neurons from a petabyte-scale super-resolution fluorescence dataset of the mouse brain on a single GPU within one week, without requiring exhaustive dense inference across the complete imaging volume.
The code is available at \url{https://github.com/albertding1113-spec/SkNeXt}.
\end{abstract}

\section{Introduction}

The morphology and connectivity of individual neurons provide the structural basis for information processing in the nervous system.
Individual neurons can extend projections over millimeters to centimeters, traverse multiple anatomical regions, and interact with numerous cells through specialized subcellular structures. 
Reconstructing complete neuronal morphology while resolving fine features such as dendritic spines and axonal boutons is therefore essential for linking local cellular architecture to brain-wide circuit organization. 
Advances in electron microscopy (EM) have enabled connectomic reconstruction at synaptic resolution, culminating in increasingly comprehensive wiring diagrams of relatively compact nervous systems such as \textit{Caenorhabditis elegans} and \textit{Drosophila melanogaster} \cite{cook2019connectomes, witvliet2021connectomes, dorkenwald2024neuronal}.
In parallel, developments in fluorescence microscopy, tissue expansion, and large-volume imaging have substantially increased both the spatial resolution and anatomical scale at which neuronal morphology can be visualized. 
Extending these approaches to mammalian brain volumes, however, inevitably generates terabyte- to petabyte-scale datasets, shifting a major bottleneck from image acquisition toward scalable neuronal reconstruction.

This challenge is particularly severe for long-range mammalian neurons.
Individual neurons may project across large fractions of the brain, requiring neuronal identity and connectivity to be maintained across thousands of spatially separated image blocks. 
Small local errors can accumulate along these trajectories, producing fragmentation, erroneous mergers, incorrect branch assignments, or loss of distal projections. 
Neuronal identity is fundamentally a long-range topological property, whereas most image-analysis networks operate on restricted local fields of view. 
Consequently, locally accurate voxel-level predictions may still produce globally incorrect neuronal reconstructions.

Existing connectomic pipelines illustrate the difficulty of scaling dense reconstruction. 
Flood-filling networks can achieve highly accurate neuronal instance segmentation by iteratively extending individual objects, but at substantial computational cost \cite{januszewski2018floodfilling}. 
Reconstruction of the adult \textit{Drosophila} brain similarly combined automated segmentation with extensive proofreading by expert annotators and community contributors \cite{dorkenwald2024neuronal}. 
Directly extending exhaustive dense segmentation and voxel-level proofreading to mammalian whole-brain datasets would therefore impose considerably greater computational and human-resource requirements.

Whole-brain fluorescence microscopy provides a complementary strategy for studying neuronal architecture at the mammalian-brain scale.
Tissue clearing, light-sheet microscopy, fluorescence micro-optical sectioning tomography (fMOST), and related approaches, particularly when combined with sparse neuronal labeling, have enabled brain-wide visualization and reconstruction of individual neuronal projections \cite{mai2024wholebody, yi2024tesos, chakraborty2019lightsheet, li2010most}.
Unlike dense segmentation, neuronal tracing represents morphology as sparse centerline graphs, commonly stored in SWC format.
These representations efficiently encode trajectories, branching relationships, and long-range connectivity while requiring only a small fraction of the data needed for voxel-wise descriptions. 
Recent computational and deep-learning approaches have further improved automated extraction of neuronal skeletons directly from microscopy images, suggesting that global topology can be reconstructed without first generating dense segmentation of the entire image volume \cite{gou2024gapr}.

Skeletons and dense segmentations capture complementary aspects of neuronal morphology.
Skeletons efficiently represent continuity and branching topology but contain limited information about neurite boundaries and fine subcellular structures.
Dense segmentation recovers these details but is computationally expensive and locally ambiguous, particularly where neighboring processes cross or run in parallel.
We therefore reasoned that large-scale neuronal reconstruction could be reformulated as a topology-first problem: 
global neuronal connectivity is first established and corrected using a compact skeleton representation, after which detailed morphology is reconstructed under the constraint of this topology.

This formulation provides two important advantages. 
First, proofreading can be concentrated on sparse neuronal trees rather than distributed throughout dense voxel volumes. 
Once corrected, the skeleton provides a persistent representation of neuronal identity that can guide reconstruction across independently processed image regions.
Second, the skeleton can serve as a spatial index for selective data access. 
Because individual neurons occupy only a small fraction of a whole-brain volume, high-resolution image data need only be retrieved from regions surrounding neuronal trajectories. 
Related systems such as UltraTracer have demonstrated that selective subvolume access can extend tracing to datasets too large to process monolithically \cite{peng2017ultratracer}.
Extending this concept from tracing to dense reconstruction offers a route to reducing both data movement and computation.

Based on these principles, we developed SkNeXt, a topology-first framework for scalable neuronal reconstruction from large volumetric microscopy datasets.
SkNeXt first reconstructs neuronal skeletons from reduced-resolution image data, enabling long-range morphology and connectivity to be established without exhaustive analysis of the original-resolution volume. 
The predicted skeletons are then proofread to correct topological errors and subsequently used as structural priors for high-resolution reconstruction, preserving neuronal identity across spatially separated regions while supporting recovery of detailed morphology and semantic features.

Crucially, SkNeXt also uses corrected skeletons to selectively retrieve high-resolution image regions along neuronal trajectories while bypassing most background and signal-free volumes.
This coupling of topology with data access allows reconstruction cost to scale more closely with neuronal morphology than with total imaging volume.
We applied SkNeXt to petabyte-scale super-resolution fluorescence imaging data of the mouse brain and demonstrated neuronal reconstruction using a single GPU. 
Together, these features provide a scalable framework for reconstructing long-range neuronal morphology from datasets for which exhaustive dense processing is increasingly impractical.

\begin{figure}[htbp]
    \centering
    \includegraphics[width=\textwidth]{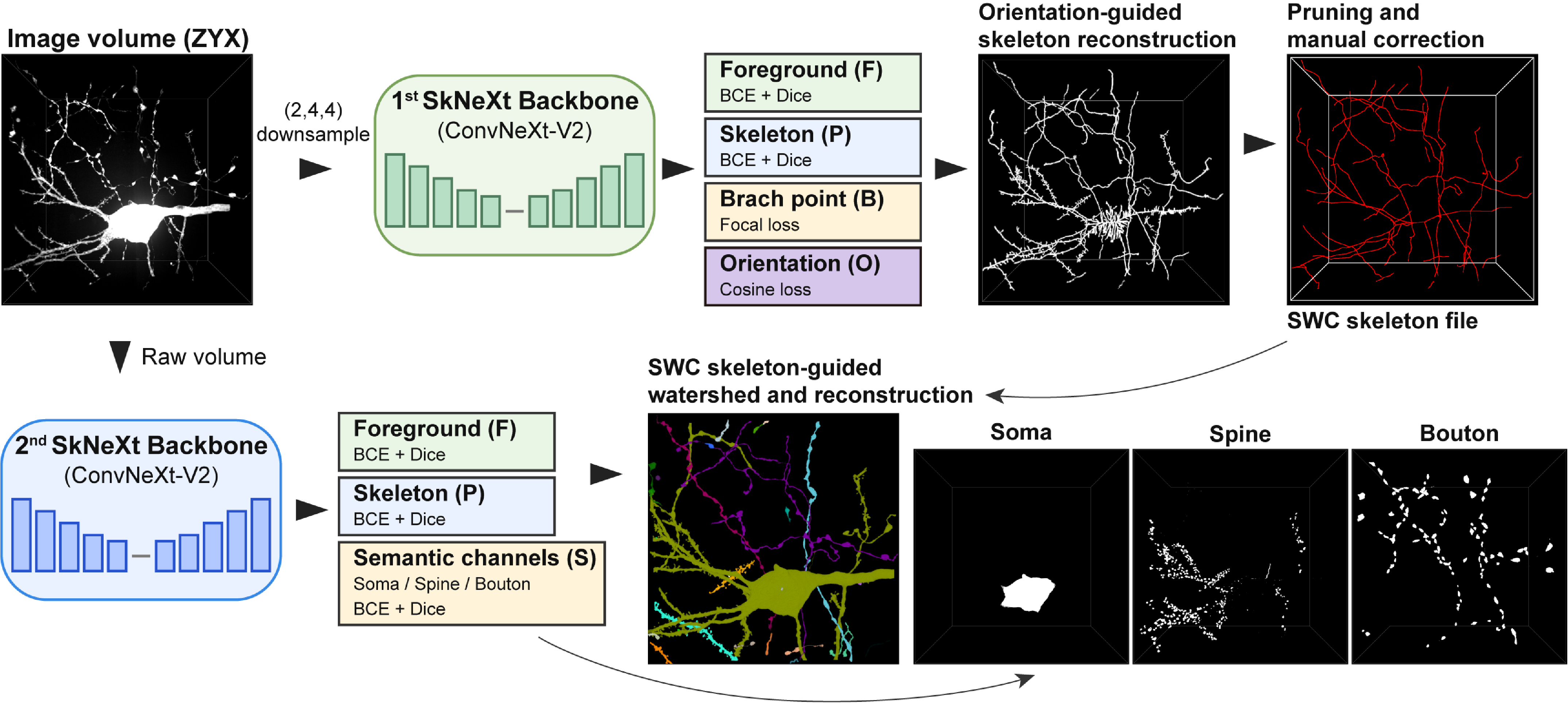}
    \caption{Overview of the SkNeXt framework.}
    \label{fig:sknext}
\end{figure}

\section{Method}
\subsection{Datasets preparation}
We used super-resolution fluorescence microscopy datasets of densely labeled neurons in the mouse brain.
To capture the morphological and imaging variability encountered during whole-brain reconstruction, we selected representative volumes from multiple brain regions containing neuronal somata, dendritic arbors, local axonal branches, and long-range projecting axons.
Regions with different neuronal densities, signal intensities, background levels, and degrees of neurite overlap were included to improve the robustness of the trained models.
Neuronal structures within the selected volumes were manually annotated using VAST.
Individual neurons were assigned unique instance labels, enabling separation of spatially adjacent or intersecting neuronal processes.
The annotated volumes were divided into training and validation datasets using a 9:1 ratio.

\subsection{Automatic SWC skeleton reconstruction}
Automatic neuronal tracing was performed using a 3D ConvNeXt V2-based multi-task network \cite{woo2023convnextv2}.
To reduce computational and I/O requirements while retaining long-range neuronal trajectories, the input microscopy volumes were downsampled by factors of 2, 4, and 4 along the $Z$, $Y$, and $X$ axes, respectively.
The network generated four prediction heads: foreground probability $F$, skeleton probability $P$, local skeleton orientation $O$, and branch-point probability $B$.
\subsubsection{Multi-task loss}
The network was optimized using a weighted multi-task objective,

\begin{equation}
\mathcal{L}
=
\lambda_F\mathcal{L}_F
+
\lambda_P\mathcal{L}_P
+
\lambda_B\mathcal{L}_B
+
\lambda_O\mathcal{L}_O,
\end{equation}

Foreground and skeleton predictions were optimized using a combination of binary cross-entropy and Dice losses,

\begin{equation}
\mathcal{L}_{\mathrm{seg}}(Q,Y)
=
\mathcal{L}_{\mathrm{BCE}}(Q,Y)
+
\lambda_D
\left[
1-
\frac{
2\sum_x Q(x)Y(x)+\epsilon
}{
\sum_x Q(x)+\sum_x Y(x)+\epsilon
}
\right],
\end{equation}

where $Q$ denotes the predicted probability map and $Y$ denotes the corresponding binary target. Thus,

Because branch points occupy only a small fraction of the image volume, focal loss was used for branch-point prediction,

\begin{equation}
\begin{aligned}
\mathcal{L}_B
=
-\sum_x
&\alpha
\left(1-B(x)\right)^{\gamma}
Y_B(x)
\log B(x)
\\
&-
(1-\alpha)
B(x)^{\gamma}
\left[1-Y_B(x)\right]
\log\left[1-B(x)\right].
\end{aligned}
\end{equation}

For orientation prediction, the output vector was first normalized. 
Then a sign-invariant cosine loss was then calculated over valid non-branching skeleton locations $M_O$,

\begin{equation}
\mathcal{L}_O
=
\frac{1}{|M_O|}
\sum_{x\in M_O}
\left[
1-
\left|
\hat{\mathbf{o}}(x)
\cdot
\mathbf{o}^{*}(x)
\right|
\right].
\end{equation}

The absolute inner product makes the loss invariant to reversal of the orientation vector.

\subsubsection{Orientation-guided skeleton reconstruction}

During inference, predictions from overlapping image blocks were merged to generate continuous foreground ($F$), skeleton ($P$), orientation ($O$), and branch-point ($B$) maps.
Candidate centerline points were extracted from high-probability regions of $P$ within the predicted foreground mask, and local non-maximum suppression was applied to obtain sparse skeleton candidates. 
Local maxima in $B$ were additionally retained as candidate branch points.

The candidate points were represented as a graph $G=(V,E)$, in which nearby nodes were connected when the intervening image region was sufficiently supported by the foreground and skeleton predictions.
For each candidate edge $(i,j)$, the connection cost combined skeleton confidence, foreground confidence, and agreement between the edge direction and the predicted local orientation:

\begin{equation}
c_{ij}
=
l_{ij}
\left[
\alpha\left(-\log(\bar{P}_{ij}+\epsilon)\right)
+
\beta\left(-\log(\bar{F}_{ij}+\epsilon)\right)
+
\gamma\left(1-\left|\bar{\mathbf{o}}_{ij}\cdot\mathbf{d}_{ij}\right|\right)
\right],
\end{equation}

where $l_{ij}$ is the physical distance between nodes, $\bar{P}_{ij}$ and $\bar{F}_{ij}$ are the mean skeleton and foreground probabilities along the connection, $\bar{\mathbf{o}}_{ij}$ is the mean predicted orientation, and $\mathbf{d}_{ij}$ is the edge direction.
Thus, high-confidence and orientation-consistent connections receive lower costs.

Tracing was initiated from a soma or predefined root node, and a minimum-cost tree was constructed over the graph.
A limited connection distance was allowed to bridge short discontinuities in the predicted skeleton when supported by foreground probability and consistent orientation.
Predicted branch-point probabilities were used to favor bifurcations at high-confidence branch locations and suppress spurious branching elsewhere.

\subsubsection{Tree pruning and SWC generation}

The initial graph was reconstructed using permissive thresholds to preserve neurite recall.
Spurious terminal branches were subsequently removed according to branch length and prediction confidence, while long or high-confidence branches were retained.
Redundant nodes along approximately straight segments were removed, and the remaining skeleton was resampled at approximately uniform spatial intervals.
Tree connectivity was then converted into parent--child relationships and exported in SWC format.

\subsection{Skeleton proofreading}
Automatically reconstructed SWC skeletons were loaded into Lychnis together with the corresponding original-resolution fluorescence images for manual inspection and correction.
Proofreading focused on topological errors introduced during automatic reconstruction.
False-positive or over-connected branches were removed, disconnected neurite segments were reconnected when supported by the underlying fluorescence signal, and incorrect parent--child relationships were corrected where necessary. 
After topology correction, skeleton nodes or branches were assigned morphological identities, including soma, axon, and dendrite, according to their anatomical continuity and local morphology.

Because proofreading was performed on sparse skeleton graphs rather than dense voxel-level segmentations, manual correction involved editing a relatively small number of nodes and edges instead of inspecting and modifying large three-dimensional label volumes. 
This substantially reduced the amount of manual interaction required for reconstruction.

\subsection{Semantic and instance segmentation}

High-resolution neuronal reconstruction was performed using a second 3D ConvNeXt V2-based multi-task network operating on the original-resolution microscopy data.
The network predicted foreground probability ($F$), skeleton probability ($P$), and semantic probability maps for neuronal structures, including soma, dendrite, and presynaptic bouton.
All outputs were treated as independent binary prediction channels.

\subsubsection{Multi-task loss}

Both the foreground/skeleton outputs and semantic outputs were optimized using a combination of binary cross-entropy (BCE) loss and Dice loss. 
For a predicted probability map $Q$ and its corresponding binary target $Y$, the segmentation loss was defined as

\begin{equation}
\mathcal{L}_{\mathrm{seg}}(Q,Y)
=
\mathcal{L}_{\mathrm{BCE}}(Q,Y)
+
\lambda_D \mathcal{L}_{\mathrm{Dice}}(Q,Y),
\end{equation}

where

\begin{equation}
\mathcal{L}_{\mathrm{Dice}}(Q,Y)
=
1-
\frac{
2\sum_x Q(x)Y(x)+\epsilon
}{
\sum_x Q(x)+\sum_x Y(x)+\epsilon
}.
\end{equation}

The foreground and skeleton losses were calculated as

\begin{equation}
\mathcal{L}_{F}
=
\mathcal{L}_{\mathrm{seg}}(F,Y_F),
\qquad
\mathcal{L}_{P}
=
\mathcal{L}_{\mathrm{seg}}(P,Y_P).
\end{equation}

For the semantic outputs, the loss was calculated independently for each semantic class and averaged across channels,

\begin{equation}
\mathcal{L}_{S}
=
\frac{1}{N_S}
\sum_{c=1}^{N_S}
\mathcal{L}_{\mathrm{seg}}(S_c,Y_{S_c}),
\end{equation}

where $N_S$ is the number of semantic classes. The total training objective was

\begin{equation}
\mathcal{L}
=
\lambda_F\mathcal{L}_{F}
+
\lambda_P\mathcal{L}_{P}
+
\lambda_S\mathcal{L}_{S},
\end{equation}

where $\lambda_F$, $\lambda_P$, and $\lambda_S$ control the relative contributions of the corresponding prediction tasks.

\subsubsection{Skeleton-guided selective inference}

For inference on terabyte- to petabyte-scale datasets, the proofread SWC skeletons were used to determine which regions of the original-resolution image volume required high-resolution processing.
The dataset was partitioned into reconstruction blocks of size $D \times H \times W$. 
For each reconstruction block, an expanded region of size $D' \times H' \times W'$ was generated by adding a safety margin along all spatial dimensions.
The expanded region provided additional image context and reduced the risk of truncating thick structures, particularly somata and proximal dendrites, near block boundaries.
Each $D' \times H' \times W'$ region was tested for intersection with the proofread SWC skeletons.
Regions containing no skeleton trajectory were skipped without loading the corresponding high-resolution image data. Only regions intersecting one or more skeletons were retrieved and passed through the network.

Network inference was performed on the expanded regions, whereas predictions corresponding to the central $D \times H \times W$ region were retained for final reconstruction.
This strategy reduced boundary artifacts while allowing high-resolution inference to be restricted to image regions surrounding reconstructed neuronal trajectories. 
Consequently, large background and signal-free regions of the dataset were bypassed without dense processing.

\subsubsection{High-resolution skeleton refinement}

The proofread SWC skeletons were initially reconstructed from downsampled image data and therefore did not necessarily coincide exactly with neuronal centerlines at the original resolution. 
To refine their positions, the predicted skeleton probability map $P$ was thresholded to obtain high-confidence centerline candidates within each processed high-resolution block.

The corresponding proofread SWC segments were rasterized into image space and used to establish connectivity between predicted skeleton regions. 
Nearby high-probability voxels in $P$ were associated with the corresponding SWC trajectory while preserving the topology and neuronal identity of the proofread skeleton. 
The resulting centerlines constituted high-resolution skeleton representations that retained the unique identity of each SWC tree.

\subsubsection{Skeleton-guided instance reconstruction}

High-resolution neuronal instances were reconstructed using marker-controlled watershed. The foreground probability map $F$ was used to define the region in which neuronal instances were allowed to grow,
$M_F=\left\{x \mid F(x) > \tau_F\right\},$where $\tau_F$ denotes the foreground probability threshold.
The watershed topographic surface was defined from the inverse foreground probability, $
T(x)=1-F(x)$, such that high-confidence neuronal regions corresponded to low-cost basins. 
The refined high-resolution skeletons were converted into a marker image, with each skeleton assigned the unique identity of its corresponding proofread SWC tree.

Marker-controlled watershed was then performed on $T(x)$ within the foreground mask $M_F$. 
Each neuronal instance expanded outward from its skeleton seed through high-confidence foreground regions until reaching background or competing instance boundaries. 
In this manner, local voxel-level predictions were assigned directly to globally defined neuronal identities.

For blocks containing multiple neurons, the unique SWC identifiers were retained throughout watershed reconstruction, allowing neighboring neuronal processes to be separated while maintaining long-range identity across independently processed blocks. 
Predictions from adjacent blocks were subsequently merged according to their shared SWC identifiers.

Finally, semantic predictions were associated with the reconstructed neuronal instances, enabling soma, dendritic, and presynaptic bouton information to be mapped onto individual neurons. 
The resulting instance and semantic reconstructions were combined with the proofread SWC skeletons to generate the final high-resolution neuronal morphology.

\subsection{Implementation details}

We applied the complete SkNeXt reconstruction workflow to multiple super-resolution neuronal imaging datasets ranging in size from approximately 0.2 to 2 PB.
All neural network training and inference experiments were performed on a workstation equipped with an NVIDIA RTX PRO 6000 Blackwell GPU.
To enable efficient random access to terabyte- and petabyte-scale image volumes, raw microscopy datasets were converted into chunked storage formats, including Imaris and OME-Zarr \cite{moore2023omezarr}.
During large-scale inference, only image chunks required by the reconstruction workflow were loaded into memory, thereby avoiding sequential access to the entire dataset.
All predicted volumetric outputs, including probability maps, semantic segmentation maps, and instance labels, were stored in OME-Zarr format. 
Zstandard (Zstd) compression was used to reduce storage requirements and I/O overhead, which was particularly important for large integer-valued instance-label volumes.

Both the skeleton reconstruction network and the subsequent semantic and instance segmentation network were implemented using a ConvNeXt-V2-based encoder--decoder architecture.
Unless otherwise specified, the input patch size was $32 \times 256 \times 256$ voxels in the $Z$, $Y$, and $X$ dimensions, respectively.
The numbers of feature channels at successive network stages were $[32, 64, 128, 256, 320]$.
The same backbone configuration was used for the two reconstruction stages, while their output heads were adapted to their respective prediction tasks.
Networks were optimized using AdamW with a maximum learning rate of $1\times10^{-4}$.
The learning rate was controlled using a warm-up cosine-decay schedule, in which an initial warm-up phase was followed by cosine annealing over the remaining training iterations.
Training each network required approximately 10 days on a single NVIDIA RTX PRO 6000 Blackwell GPU.
Following initial skeleton inference, the automatically reconstructed SWC files were manually proofread to correct residual topological errors, including incorrectly connected or disconnected branches.
Proofreading required approximately 0.5 person-hour per neuronal skeleton.
The corrected SWC skeletons were subsequently used to guide high-resolution semantic and instance segmentation.
Depending on dataset size and neuronal density, inference of the final dense segmentation results required approximately 3--14 days per dataset on a single GPU.
Together, the skeleton-guided selective data-access strategy substantially reduced the amount of high-resolution image data that required neural network inference, enabling the reconstruction workflow to be applied to datasets approaching the petabyte scale with workstation-level computational resources.

\section{Results}

\begin{figure}[htbp]
    \centering
    \includegraphics[width=0.8\textwidth]{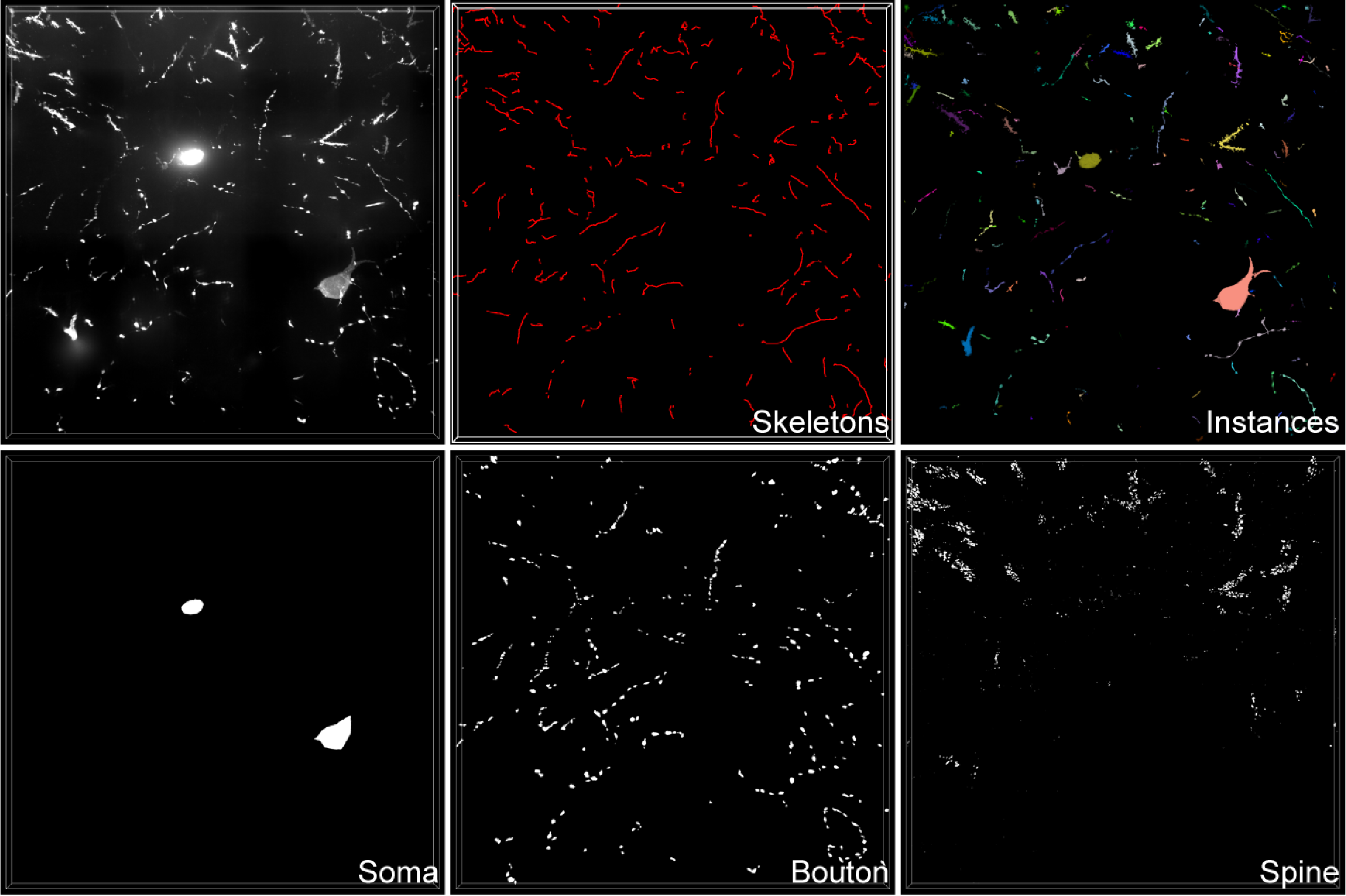}
    \caption{Segmentation results of a single view.}
    \label{fig:SingleViewSegmentation}
\end{figure}

We first evaluated the complete SkNeXt workflow on a single-view super-resolution dataset with a volume size of $40 \times 4096 \times 4096$ voxels.
Neuronal skeletons were directly reconstructed from the raw fluorescence images and required only minimal manual proofreading (Fig.~\ref{fig:SingleViewSegmentation})
The reconstructed skeletons preserved both major neuronal branches and long, thin axonal processes, providing a topological representation of individual neurons prior to dense instance segmentation.
Conventional watershed-based instance segmentation is prone to under-segmentation in densely labeled neuronal regions, where adjacent somata, dendrites, and axons frequently overlap or remain connected in the foreground prediction. 
To address this problem, SkNeXt uses the reconstructed neuronal skeletons as identity-aware seeds for watershed segmentation. 
Because each skeleton represents an individual neuron and extends along its neuronal processes, the skeleton seeds spatially separate neighboring neurons in densely labeled regions while simultaneously maintaining the continuity of thin axons that are otherwise susceptible to fragmentation and over-segmentation.
Consequently, each reconstructed neuronal instance is explicitly associated with a corresponding skeleton identity, and the number of segmented neuronal instances is constrained to match the number of input skeletons.
This one-to-one correspondence between skeletons and neuronal instances substantially simplifies the subsequent proofreading procedure.
In contrast to conventional instance-segmentation workflows, which often require extensive manual merging of fragmented neuronal processes and splitting of incorrectly fused neurons, skeleton-guided reconstruction largely eliminates these operations. 
Manual proofreading can therefore focus primarily on correcting residual local segmentation errors rather than reconstructing neuronal identities, substantially reducing the labor and time required for post-processing.
In parallel with neuronal instance reconstruction, SkNeXt simultaneously predicts semantic labels for neuronal somata and subcellular structures. 
These semantic predictions are subsequently assigned to their corresponding skeleton identities through the reconstructed neuronal instances, enabling each soma and subcellular structure to be associated with a specific neuron.
The resulting identity-aware semantic annotations provide a unified representation of neuronal morphology and subcellular organization, facilitating downstream analyses of structure distribution, neuronal morphology, and neuron-specific spatial organization.

\begin{figure}[htbp]
    \centering
    \includegraphics[width=\textwidth]{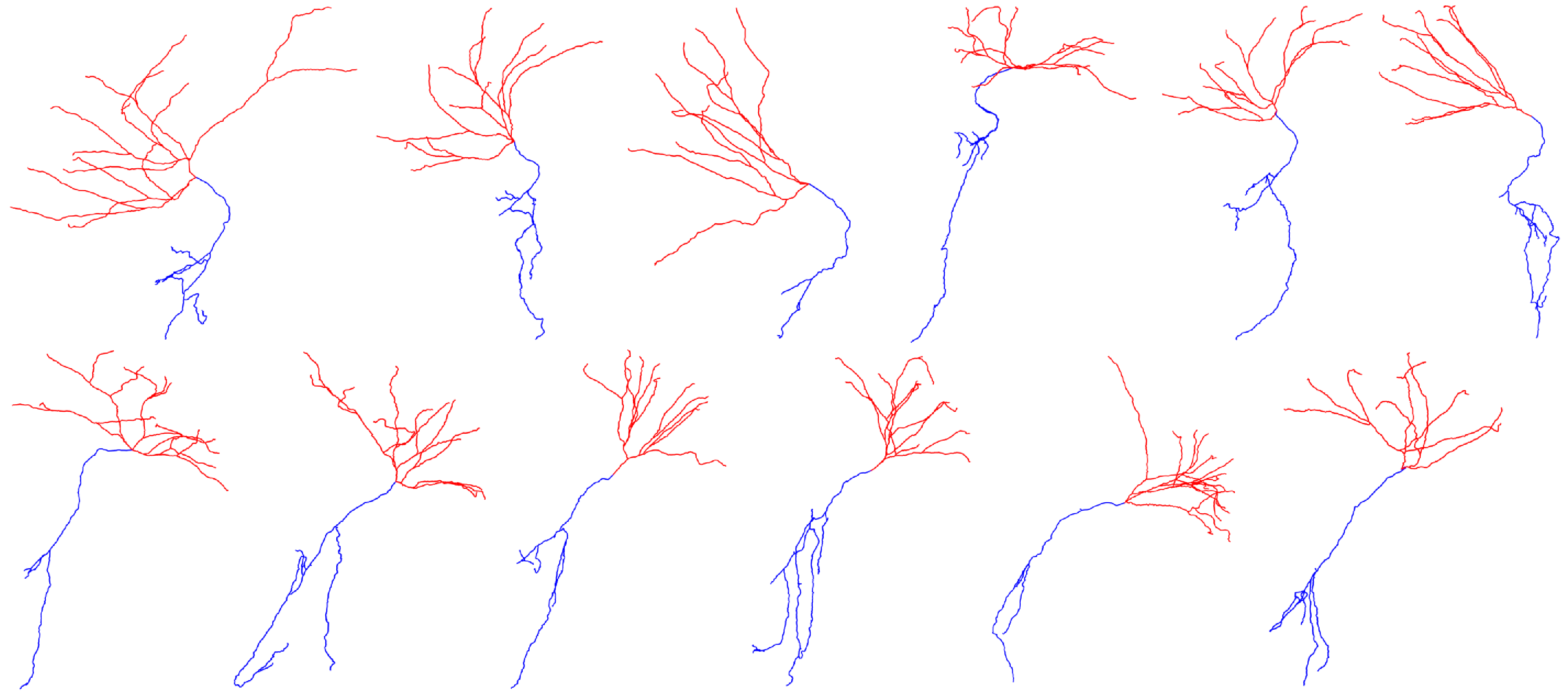}
    \caption{Skeleton reconstruction of hippocampus neurons.}
    \label{fig:SkeletonReconstrcution}
\end{figure}

We next applied SkNeXt to a substantially larger, 0.3-PB super-resolution hippocampal neuronal dataset to evaluate its performance at scale.
Initial neuronal skeletons were automatically reconstructed by SkNeXt and subsequently manually corrected and annotated with axonal and dendritic identities (Fig.~\ref{fig:SkeletonReconstrcution}).
In contrast to the single-view dataset, complete elimination of manual proofreading was not feasible at this scale and labeling density.
The hippocampal dataset contained extensively interwoven neurites, frequent crossings between neighboring neuronal processes, and weak or extremely thin axonal segments, which occasionally introduced incorrect connections or discontinuities during automatic skeleton reconstruction.
Importantly, however, proofreading the automatically reconstructed skeletons was substantially more efficient than reconstructing neurons directly from the raw image volume.
On average, proofreading a SkNeXt-generated skeleton required approximately $0.5$ person-hours per neuron, whereas direct manual reconstruction from the raw dataset required approximately $4$ person-hours per neuron.
Thus, the skeleton-first workflow reduced the manual reconstruction workload by approximately eightfold.
Moreover, proofreading was performed on a sparse topological representation rather than repeatedly navigating through the full-resolution image volume, allowing erroneous branches, missing connections, and neurite identities to be corrected more efficiently.

Following skeleton proofreading, we performed skeleton-guided instance and semantic segmentation over the entire 0.3-PB dataset.
Despite the scale of the input data, the complete segmentation inference was finished in approximately three days using a single NVIDIA RTX PRO 6000 Blackwell GPU.
This demonstrates that the skeleton-guided selective inference strategy can substantially reduce the computational burden of dense reconstruction by restricting full-resolution processing to regions associated with neuronal trajectories rather than exhaustively processing the entire imaging volume.

\begin{figure}[htbp]
    \centering
    \includegraphics[width=\textwidth]{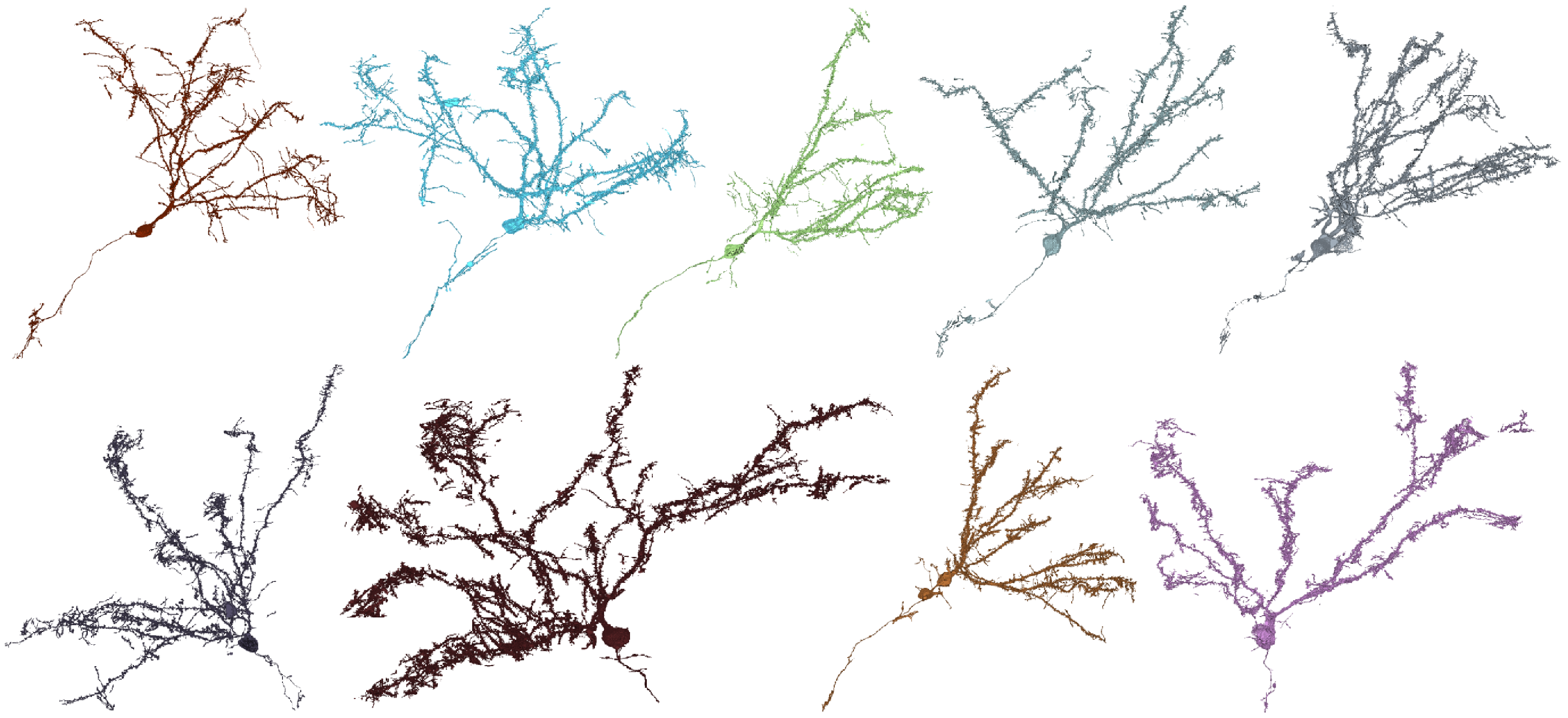}
    \caption{Instance segmentation of hippocampus neurons.}
    \label{fig:InstanceSegmentation}
\end{figure}

The resulting instance segmentation preserved neuronal identity from the corrected skeletons throughout the reconstructed volume (Fig.~\ref{fig:InstanceSegmentation}).
Somata and neuronal processes were continuously reconstructed along the corresponding skeleton trajectories, with no apparent merging of neighboring neurons or fragmentation of individual processes along the examined skeleton paths.
In particular, thin axonal segments that are susceptible to over-segmentation in conventional patch-wise approaches remained associated with their parent neurons, whereas closely juxtaposed neuronal processes were separated according to their respective skeleton identities.
Semantic predictions further resolved neuron-associated subcellular structures, including neuronal somata, presynaptic boutons along axons, and postsynaptic structures along dendrites.
Because these semantic labels were integrated with skeleton-guided instance reconstruction, individual subcellular structures could be directly assigned to their parent neurons and to the corresponding axonal or dendritic compartments.
Together, these results demonstrate that SkNeXt enables topology-preserving neuronal reconstruction and identity-aware subcellular annotation across petabyte-scale super-resolution microscopy datasets while substantially reducing both computational and manual reconstruction costs.

\section{Conclusion}
Here, we introduce SkNeXt, a deep-learning-based workflow for topology-guided neuronal reconstruction from large-scale super-resolution fluorescence microscopy datasets.
SkNeXt first reconstructs neuronal skeletons directly from raw images and then uses manually proofread skeletons as topological priors for subsequent instance and semantic segmentation.
By separating topology reconstruction from dense voxel-level segmentation, the workflow preserves neuronal identity across long and complex processes while reducing common segmentation errors such as fragmentation of thin axons and merging of densely interwoven neurites.

A major advantage of SkNeXt is that manual intervention is shifted from labor-intensive voxel-level reconstruction to proofreading of sparse SWC skeletons.
In our large-scale hippocampal dataset, proofreading automatically reconstructed skeletons required approximately $0.5$ person-hours per neuron, compared with approximately $4$ person-hours for direct manual reconstruction from raw images. 
The corrected skeletons further provide identity-aware seeds for dense segmentation, enabling reconstructed voxels and semantic structures to be consistently assigned to individual neurons without extensive post hoc merging or splitting.

SkNeXt is also designed for petabyte-scale datasets.
By using skeleton-guided selective data access, full-resolution inference is restricted to image regions traversed by neuronal processes, substantially reducing unnecessary computation over background regions.
Using a single GPU, SkNeXt completed instance and semantic segmentation of a 0.3-PB hippocampal dataset within approximately three days.
In addition to reconstructing complete neuronal morphology, the workflow simultaneously identifies neuronal somata and subcellular structures, including axonal boutons and dendritic spines, and associates these structures with their corresponding neuronal identities.

Although manual proofreading remains necessary in regions containing highly interwoven neurites, weak signals, or extremely thin neuronal processes, SkNeXt substantially reduces the amount of human labor required for large-scale neuronal reconstruction.
Overall, SkNeXt provides a scalable framework that integrates skeleton reconstruction, topology-guided instance segmentation, and neuron-aware semantic annotation, bridging the gap between petabyte-scale fluorescence imaging and quantitative analysis of neuronal morphology and subcellular organization.

\bibliographystyle{unsrtnat}
\bibliography{references}

\end{document}